\PassOptionsToPackage{unicode}{hyperref}
\PassOptionsToPackage{hyphens}{url}
\documentclass[
]{article}
\usepackage{lmodern}
\usepackage{amssymb,amsmath}
\usepackage{ifxetex,ifluatex}
\ifnum 0\ifxetex 1\fi\ifluatex 1\fi=0 % if pdftex
  \usepackage[T1]{fontenc}
  \usepackage[utf8]{inputenc}
  \usepackage{textcomp} % provide euro and other symbols
\else % if luatex or xetex
  \usepackage{unicode-math}
  \defaultfontfeatures{Scale=MatchLowercase}
  \defaultfontfeatures[\rmfamily]{Ligatures=TeX,Scale=1}
\fi
\IfFileExists{upquote.sty}{\usepackage{upquote}}{}
\IfFileExists{microtype.sty}{% use microtype if available
  \usepackage[]{microtype}
  \UseMicrotypeSet[protrusion]{basicmath} % disable protrusion for tt fonts
}{}
\makeatletter
\@ifundefined{KOMAClassName}{% if non-KOMA class
  \IfFileExists{parskip.sty}{%
    \usepackage{parskip}
  }{% else
    \setlength{\parindent}{0pt}
    \setlength{\parskip}{6pt plus 2pt minus 1pt}}
}{% if KOMA class
  \KOMAoptions{parskip=half}}
\makeatother
\usepackage{xcolor}
\IfFileExists{xurl.sty}{\usepackage{xurl}}{} % add URL line breaks if available
\IfFileExists{bookmark.sty}{\usepackage{bookmark}}{\usepackage{hyperref}}
\hypersetup{
  pdftitle={Sample Selection Bias in Evaluation of Prediction Performance of Causal Models},
  pdfauthor={James P. Long and Min Jin Ha},
  hidelinks,
  pdfcreator={LaTeX via pandoc}}
\usepackage[margin=1in]{geometry}
\usepackage{graphicx,grffile}
\makeatletter
\def\maxwidth{\ifdim\Gin@nat@width>\linewidth\linewidth\else\Gin@nat@width\fi}
\def\maxheight{\ifdim\Gin@nat@height>\textheight\textheight\else\Gin@nat@height\fi}
\makeatother
\setkeys{Gin}{width=\maxwidth,height=\maxheight,keepaspectratio}
\makeatletter
\def\fps@figure{htbp}
\makeatother
\usepackage{xcolor}

\usepackage{enumitem}
\usepackage{natbib}
\usepackage{booktabs}

\newcommand{\rev}[1]{\textcolor{black}{#1}}

\title{Sample Selection Bias in Evaluation of Prediction Performance of Causal Models}
\author{James P. Long\thanks{jplong@mdanderson.org} \, \, \, \, \,  Min Jin Ha\\ \\Department of Biostatistics\\ University of Texas MD Anderson Cancer Center}
\date{\today}

\begin{document}
\maketitle

{
\setcounter{tocdepth}{2}
\tableofcontents
}

\hypertarget{abs}{%
\section*{Abstract}\label{abs}}

Causal models are notoriously difficult to validate because they make untestable assumptions regarding confounding. New scientific experiments offer the possibility of evaluating causal models using prediction performance. Prediction performance measures are typically robust to violations in causal assumptions. However prediction performance does depend on the selection of training and test sets. In particular biased training sets can lead to optimistic assessments of model performance. In this work, we revisit the prediction performance of several recently proposed causal models tested on a genetic perturbation data set of Kemmeren \citep{kemmeren2014large}. We find that sample selection bias is likely a key driver of model performance. We propose using a less-biased evaluation set for assessing prediction performance and compare models on this new set. In this setting, the causal models have similar or worse performance compared to standard association based estimators such as Lasso. Finally we compare the performance of causal estimators in simulation studies which reproduce the Kemmeren structure of genetic knockout experiments but without any sample selection bias. These results provide an improved understanding of the performance of several causal models and offer guidance on how future studies should use Kemmeren.

%\keywords{causal inference, prediction, sample selection bias, genetic perturbation experiments}

%\footnotetext{\textbf{Abbreviations:} ANA, anti-nuclear antibodies; APC, antigen-presenting cells; IRF, interferon regulatory factor}

\section{Introduction}\label{intro}

Modern scientific experiments offer the possibility to evaluate causal models using prediction performance. A causal estimator predicts whether $X$ is a cause of $Y$ for a large number of $(X,Y)$ pairs. Ground truth for some subset of these pairs is obtained and is used to score the predictions. Prediction performance assesses models without relying on statistical assumptions regarding the joint distribution of the random variables or complex and often unverifiable assumptions regarding confounding.

Large scale gene perturbation experiments offer one such opportunity. In these experiments, the expression of certain genes is perturbed and the effects of these perturbations on other genes is measured. These experiments provide insight into which genes causally influence each other. For gene perturbation experiments, causal prediction performance may be assessed as follows. Gene X has a causal effect on gene Y if when gene X is set to a value outside its ``normal'' range, the expression of Y is outside its ``normal'' range. Given $p$ genes there are $p(p-1)$ possible ordered cause--effect pairs (excluding self pairs). Ground truth can be obtained for some fraction of these gene pairs via knockout or knockdown experiments where the expression of a gene is set to a very low value. A causal model may use observational and/or perturbation data to make causal predictions for all gene pairs. The ground truth from the knockout experiments can then be used to score models and compare performance across models. High scoring models could be used to prioritize limited resources for follow--up experiments.

This strategy was used to compare performance of several recently proposed causal models on a yeast knockout data set produced in \cite{kemmeren2014large} (hereafter Kemmeren). The causal models assessed include Invariant Causal Prediction (ICP) and the Causal Dantzig (CD) which fit regression models which are invariant across data collection environments \citep{rothenhausler2019,peters2016causal,meinshausen2016methods} and Local Causal Discovery (LCD) which tests for dependence among gene expression and context variables \citep{versteeg2019boosting}. These models were compared to standard (non--causal) models such as Lasso regression. The causal models (ICP, CD, LCD) are expected to outperform non--causal models because when $X$ and $Y$ are correlated, non--causal models are unable to distinguish between $X$ causes $Y$, $Y$ causes $X$, or no causal relationship (correlation induced by a third variable which is a cause of both $X$ and $Y$). Indeed on the Kemmeren data set ICP, CD, and LCD were all found to have performance superior to non--causal models (in particular the Lasso and Boosting).

\rev{The purpose of this work is to identify a source of bias in performance evaluation on Kemmeren, propose an alternative, less-biased method of evaluation, and study the performance of the causal estimators in simulations which reproduce the structure of the Kemmeren data.} These results provide an improved understanding of the predictive performance of these causal models on gene perturbation data sets. Our contributions and results are summarized as follows: The essential problem in Kemmeren is that perturbations (knockouts) are performed not on a random sample of genes but on genes which are believed to be causes of expression changes in other genes (``putative regulators'' according to \cite{kemmeren2014large}). \rev{Sample selection bias produces optimistic assessments of the predictive performance of causal models because correct predictions are more likely to have ground truth available (and thus be scored) than incorrect predictions.} We see evidence of this when reviewing top ranked causal gene pairs returned by the causal estimators. To address this problem, we propose scoring only causal predictions $(X,Y)$ where ground truth is available for both the $X$ and $Y$ knockout ($(X,Y)$ and $(Y,X)$ have ground truth available). We reassess the performance of several causal estimators on Kemmeren using this new criteria. \rev{Under this new criteria, the causal estimators do not demonstrate performance improvement over Lasso regression.} To further explore these issues in an environment where biased follow--up does not effect results, we simulate data following the experimental setup of Kemmeren. \rev{These simulations reproduce the sample size, data dimension, and knockout complexity of Kemmeren.} None of the previous works studying Kemmeren presented simulation results from data generating distributions meant to approximate Kemmeren. \rev{In the simulation settings studied, causal models are unable to consistently outperform association based estimators.} The simulations provide further evidence that the experimental setup of Kemmeren is challenging for discovering cause--effect gene pairs. 

This work is organized as follows. Section \ref{problem} introduces several of the causal estimators studied in this work, describes the Kemmeren data set, and identifies a potential source of bias in performance assessment on Kemmeren. Section \ref{score} proposes a new criteria for scoring causal predictions on Kemmeren and reassesses the performance of models based on this metric. Section \ref{sim} presents simulation results which provide insight into the difficultly causal inference with the Kemmeren data set. We conclude with discussion in Section \ref{discuss}.

\section{Causal Models and Kemmeren}\label{problem}

%Idea: eventual goal of such methods would be to not have to do the experiment. examples: drug combination, crispr, and gene knockout.
%Include: Discussion of tank bias
%This approach to model performance contrasts sharply with traditional measure of performance which usually rely on 

\subsection{Causal Inference and Prediction}\label{causal-models}

Causal models seek to answer how changing a variable $X$ will effect another variable $Y$. Causal estimands may be formally expressed in the Neyman--Rubin potential--outcomes framework or via Pearl counterfactuals derived from structural equations \citep{rubin2005causal,pearl2009causal}. For example, in the Neyman--Rubin potential--outcomes framework, $\mathbb{E}[Y^{X=x}]$ is the expected value of $Y$ when $X$ is set to $x$ by an external intervention. \rev{Formally, $Y^{X=x}$ is a set of random variables (indexed by $x$) called potential outcomes which are distinct from $Y$. Causal estimands are not generally identifiable from the joint distribution of the observed random variables $(X,Y)$. Thus statistical assumptions on the joint distribution (e.g. $Y$ is a linear function of $X$) are not sufficient for causal inference. Causal estimators require a separate set of causal assumptions regarding the potential outcome random variables. For example, the ``ignorability assumption'' $Y^{X=x} \perp \!\!\! \perp X | C$ (read as ``the value $Y$ would take if $X$ is set to $x$ is independent of $X$ given $C$'') and causal consistency assumptions imply $\mathbb{E}[Y^{X=x}]=\mathbb{E}[\mathbb{E}[Y|C,X=x]]$. Since $\mathbb{E}[\mathbb{E}[Y|C,X=x]]$ is a function of the joint distribution of $(X,Y,C)$, under the ignorability and consistency assumptions the causal estimand of interest, $\mathbb{E}[Y^{X=x}]$, is identifiable. Causal assumptions such as ignorability can be especially questionable (relative to statistical assumptions) since they are generally not verifiable from observational data (even asymptotically).}

Recently several works have sought to validate causal models using prediction. This form of validation is possible when there are many causal questions of interest and ground truth can be obtained for a subset. A statement such as ``Causal model $M$ provides an accurate estimate of the causal effect of smoking on lung cancer'' cannot generally be validated using prediction performance because there is a single causal effect of interest (effect of smoking on cancer) and validating the model experimentally would obviate the need for the model in the first place (along with being ethically and financially impractical).

In contrast, a statement such as ``Causal model $M$ successfully identifies cause--effect yeast gene pairs'' can be validated in a prediction accuracy framework. Model $M$ predicts whether gene $X$ causes gene $Y$ for every $(X,Y)$ pair (e.g. ranks all gene pairs from most to least likely to be cause--effect pairs). %\rev{The data used to construct model $M$ and then rank gene pairs could purely observational where the expression of $p$ genes was measured across $n$ samples with no external interventions. Or the data could be a mixture of observation and interventional data. With interventional data, perturbations such as drugs or gene knockouts are performed which alter the expression of one or a set of genes. Such interventional data is often necessary to separate associational from causal relations.}

Ground truth is obtained for some subset of these predictions by external interventions on the purported causes. \rev{The method of obtaining ground truth may vary from application to application. One method is to ``knockout'' gene $X$. In a knockout experiment, $X$ is set to a low value (typically outside its normal range) and then $Y$ is observed. Gene $X$ has a causal effect on $Y$ if the distribution under the knockout is different from the distribution in an observational (wild type) condition, i.e. $Y^{X=0} \neq_d Y$. Since we may only have one knockout of $X$, we will only have one observation $y^{x=0}$ generated from the $Y^{X=0}$ distribution. Thus one cannot conclusively determine $Y^{X=0} \neq_d Y$. Further it may be of interest to identify gene pairs $(X,Y)$ for which $X$ has a large causal effect on $Y$, i.e. the distribution of $Y^{X=0}$ is very different from the distribution of $Y$. One possibility is to estimate the range of $Y$ from a non--interventional sample $y_1,\ldots,y_n$. If the knockout sample is outside the range of observational $Y$, i.e. $y^{x=0} \notin [\min y_i, \max y_i]$, then we conclude that $X$ has a (most likely large) effect on $Y$.}

%One could adopt a stricter definition of causal effect and require a large shift under the knockout setting, e.g. $|(\mathbb{E}[Y|do(X=0)] - \mathbb{E}[Y])/sd(Y)| > t$ where $t$ is some thre

\rev{Once ground truth, however defined, is obtained for a sample of $(X,Y)$ pairs, the performance of a set of causal models $M_j$ for $j=1,\ldots,J$ can be computed and compared. For example we may consider $M_j$ to be the best causal model if it has the highest accuracy or the largest area under the Receiver Operating Characteristic curve as compared with all other models (when scored on the set of $(X,Y)$ for which ground truth is available). If the performance of model $M_j$ is deemed sufficiently good by some measure, then performance on non--validated causal pairs $(X,Y)$ should also be good as well (up to sampling variability). In practical terms $M_j$ could be used to guide the choice of future knockout experiments to maximize the number causal pairs discovered.}

\rev{Prediction performance measures do not rely on validity of either the statistical or causal assumptions used by the model. For example,  $M_1$ may make ignorability assumptions while $M_2$ may avoid these assumptions using instrumental variable techniques but instead assume linear causal relations among the variables. Which model is better? The standard approach is to attempt to assess which model assumptions are more plausible for the given application. This is very difficult. Both models are approximations and will be wrong to some degree. The ignorability assumption of model $M_1$ is generally impossible to validate, even asymptotically. An alternative is to select the model with better causal prediction performance. This prediction based evaluation of causal models can be seen as an extension of Breiman's proposals for model validation to the field of causality \citep{breiman2001statistical}.}

\rev{A major caveat to the above discussion regards the selection of the set of $(X,Y)$ on which ground truth is obtained. If ground truth is available for a simple random sample of all potential causal pairs, then we expect that a model's performance on the ground truth set will generalize well to the set of untested pairs. However if the ground truth set represents a biased selection, e.g. highly enriched for true effects, then the performance may not generalize. We discuss this issue now in the context of the Kemmeren data set.}

%Recently the predictive performance of several causal models has been evaluated and compared on a gene knockout data set of Kemmeren. Two new methods, Invariant Causal Prediction (ICP) and the Causal Dantzig (CD) \citep{peters2016causal,rothenhausler2019}, construct estimators which are invariant across data collection environments. These environments could be an observational environment where no interventions are made and an experimental environment where external interventions (e.g. gene knockouts) are performed. ICP assumes no hidden confounding and is computationally challenging, but can handle diverse types of interventions. The CD is computationally faster than ICP and is consistent under hidden confounding. However the types of interventions permitted by CD are more limited than ICP. We refer to the respective papers for details on how the models are constructed and consistency guarantees. Our goal here is to study how these models are evaluated based on prediction performance with the Kemmeren data set.

\hypertarget{kemmeren}{%
\subsection{Kemmeren}\label{kemmeren}}

\begin{figure}[t]
  \centering
a)  \includegraphics[width=0.4\textwidth]{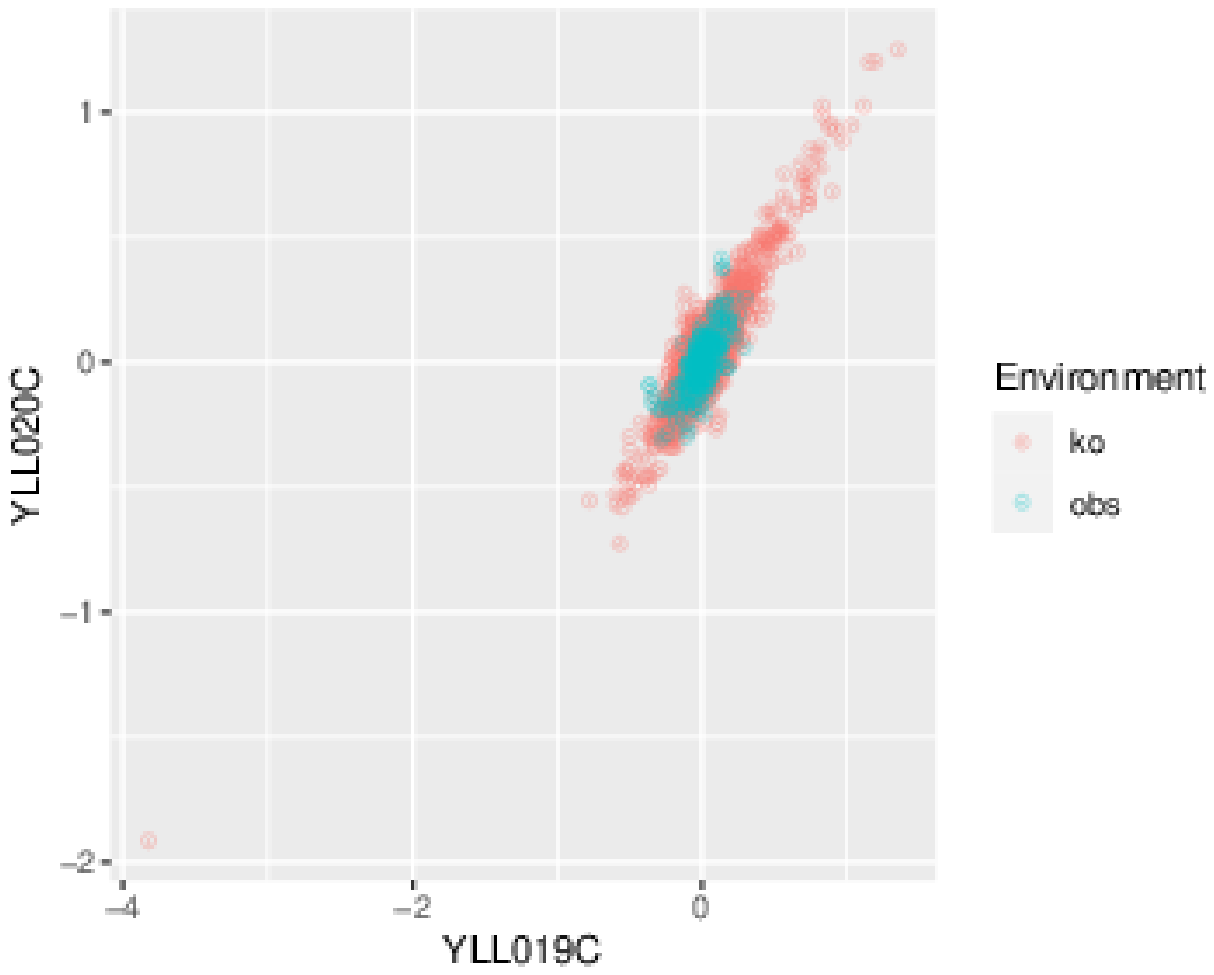}
b)  \includegraphics[width=0.4\textwidth]{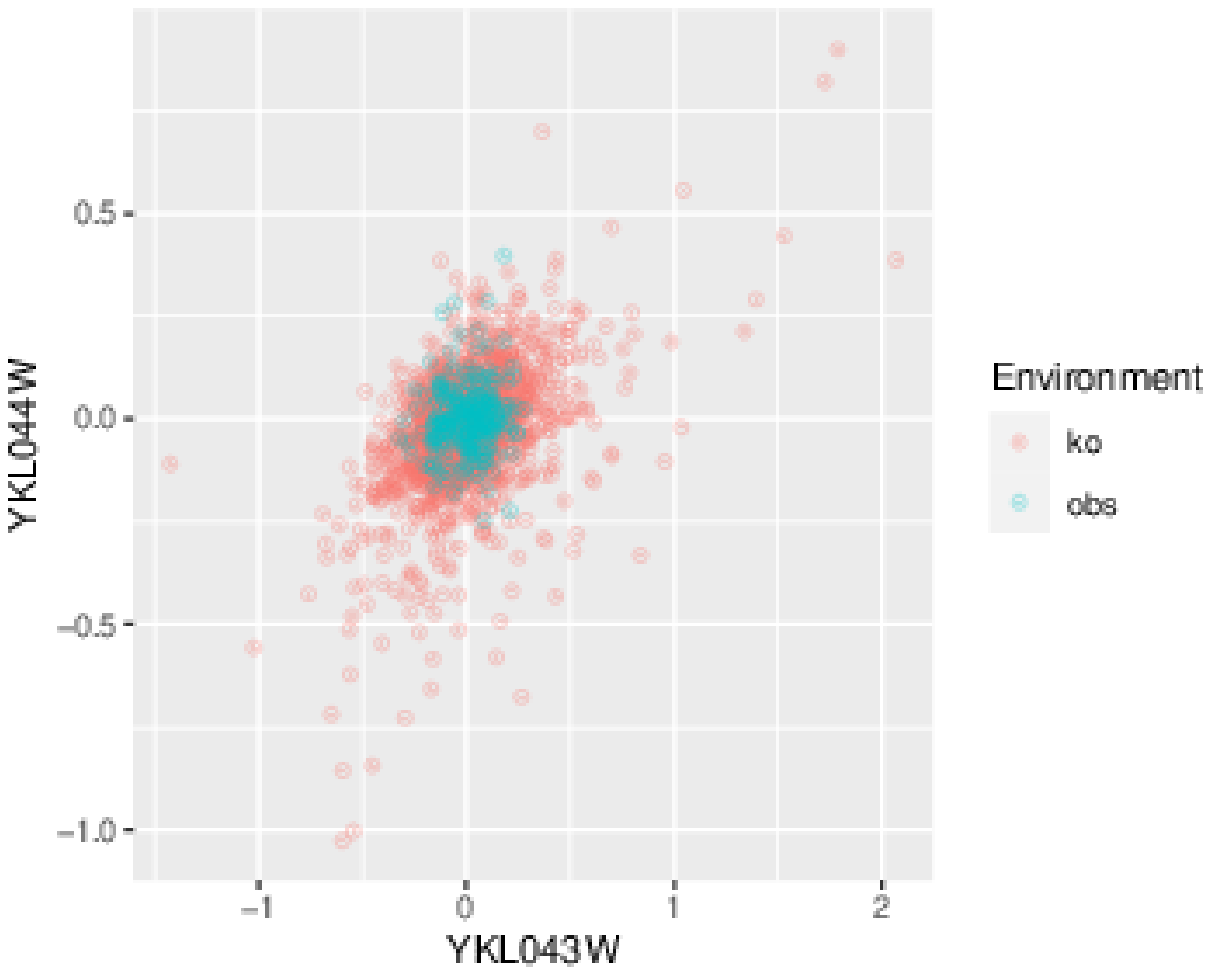}
\caption{Two gene pairs from Kemmeren. \label{fig:gene-pairs}}
\end{figure}

Kemmeren measured expression of $p=6170$ genes with $n_1 = 262$ wild-type samples where no intervention was performed and $n_2 = 1479$ interventional samples where one gene was knocked out. \footnote{Data from file \texttt{Kemmeren.RData} hosted at \url{http://deleteome.holstegelab.nl/downloads_causal_inference.php}. This data set and all code used to analyze it are available in the github repository \url{https://github.com/longjp/causal-bias-code}.} Each gene was knocked out at most once (so there are $n_2$ genes which were knocked out once and $p - n_2$ genes which were not knocked out). Figure \ref{fig:gene-pairs} shows scatterplots of the observational and interventional data for two pairs of genes. The red points (ko) are interventional samples where one gene's expression is set very low. The blue points (obs) are non-interventional samples. The variance of the gene expression is higher for the interventional samples because knockouts perturb the expression of the knocked out gene and all downstream targets, leading to greater variability in expression.

Kemmeren has been used to evaluate the predictive performance of several causal models including ICP, the CD and LCD \citep{peters2016causal,rothenhausler2019,versteeg2019boosting,meinshausen2016methods}. The performance of these models has been compared to purely association based methods such as Lasso regression. The following is a summary of how these models are fit and evaluated on Kemmeren. We follow closely the procedures used in the existing studies so that we can reproduce their results. Since each of these studies used somewhat different evaluation strategies, we do not exactly follow any single study.

Since there are $p$ genes, there are $p(p-1)$ ordered pairs of genes $(i,j)$ for which causal predictions can be made. The causal models rank all these ordered pairs from most likely to least likely to be cause--effect pairs. Ground truth for the gene pair $(i,j)$ is determined by observing the expression of gene $j$ when gene $i$ is knocked out. Following \cite{peters2016causal}, we say that gene $i$ has a \textit{significant causal effect} on gene $j$ if in the sample where gene $i$ is knocked out, the expression of gene $j$ is outside of the range of gene $j$ measured in the $n_1$ observational samples. See \cite{versteeg2019boosting} and \cite{meinshausen2016methods} for other possible definitions of causal effect.

Ground truth (i.e. knowledge of whether $i$ has a significant causal effect on $j$) is only available for $n_2(p - 1)=9,125,430$ of these gene pairs (roughly 25\% of samples). When the causal models are scored, only these predictions are considered. Predictions for the remaining $\sim 75\%$ of genes pairs are not used because ground truth is not available. About 6\% of gene pairs with ground truth are significant causal effects.

All models are fit using a cross--validation strategy which prevents the model from using the result of the $i$ knockout when making a prediction on the $(i,j)$ cause--effect pair. In particular, the $n_2$ interventional samples are split into $K$ folds. The causal models are trained on all the $n_1$ observational samples and $K-1$ folds of the interventional samples. The causal models then make predictions for the held out interventions (approximately $n_2/K \times p$ pairs). Here we set $K=3$.

Finally the models are trained on 100 bootstrap samples of the training data ($n_1$ wild type samples and $K-1$ folds of interventional data). For each bootstrap sample the models are fit and coefficients computed. Coefficients are ranked by absolute size. These ranks are averaged across the bootstrap samples to produce final rankings. This bootstrap procedure does not use the held out fold of data.

We assess the performance of four methods:
\begin{enumerate}[itemsep=0pt,parsep=0pt]
\item Lasso Regression (L1): The response gene is regressed on the expression of all other genes, ignoring knockout information. The \texttt{glmnet} package is used to fit the model \citep{friedman2010regularization}. The model with 4 nonzero coefficients is selected.
\item Causal Dantzig (CD): The CD is fit to the 4 variables selected by L1. \rev{We briefly describe the CD now and refer readers to \cite{rothenhausler2019} for full details. Let $X$ denote a set of genes that may be causes of $Y$. The CD assumes a linear model 
\begin{equation*}
Y = \beta^TX + \epsilon
\end{equation*}
where the error term $\epsilon$ may be correlated with $X$ to account for hidden confounders. The parameter $\beta_j$ from the above model is the causal effect of $X_j$ on $Y$. With hidden confounders and reverse causality, regressing $Y$ and $X$ is well known to produce inconsistent estimates of $\beta$. The CD assumes observations $(X,Y)$ are generated from environments $e \in \{1,2\}$. For Kemmeren, $e=1$ corresponds to the $n_1$ observational samples while $e=2$ corresponds to the $n_2$ knockout samples. Let $\bf{X}^e$ ($\bf{Y}^e$) represent the design matrix (response vector) for environment $e$. The CD estimates $\beta$ with
\begin{equation*}
\widehat{\beta}_{CD} = \left(\frac{1}{n_1}\bf{X}^{1^T}\bf{X}^1 - \frac{1}{n_2} \bf{X}^{2^T}\bf{X}^2\right)^{-1} \left(\frac{1}{n_1}\bf{X}^{1^T}\bf{Y}^1 - \frac{1}{n_2}\bf{X}^{2^T}\bf{Y}^2\right).
\end{equation*}
The form of the estimator $\widehat{\beta}_{CD}$ is motivated by seeking inner product invariance across environments. Under assumptions on the complexity and form of environment interventions and regularity assumptions on the error term, $\widehat{\beta}_{CD}$ is an asymptotically normal estimator of $\beta$. The CD is fit using the \texttt{causalDantzig} function in the \texttt{InvariantCausalPrediction} R package (version 0.7-1).}
\item Invariant Causal Prediction (ICP): ICP is fit to the 4 variables selected by L1 and maximin coefficients returned. \rev{We briefly describe ICP now and refer readers to \cite{peters2016causal} for full details. ICP identifies a set coefficients $R$ such that the distribution $Y^e|X_R^e=x$ does not depend on the environment $e$ for any $x$. The implementation of ICP used here (as well as in the original application of ICP to Kemmeren in \cite{peters2016causal}) assumes linear models so $Y^e = X_R^{eT}\beta_R^{e} + \epsilon^e$. The set $R$ and the coefficient values $\beta_R$ are identified by first fitting the linear model using least squares and then testing if a) the distribution of $\epsilon^e$ depends on $e$ and b) the value of $\beta_R^e$ depend on $e$. For a given $R$, if either hypothesis is rejected, then $R$ is rejected and is unlikely to be the set of causes of $Y$. If coefficient sets $\{R_k\}_{k=1}^K$ are not rejected, then the estimated causal set is $\widehat{R} = \cap_{k=1}^K R_j$. The maximin coefficient for $\beta_j$ is computed by taking a union of the $K$ confidence intervals for $\beta_j$ (one confidence interval for each $R_k$) and then selecting the point in the confidence interval union closest to $0$. ICP is fit using the \texttt{ICP} function in the \texttt{InvariantCausalPrediction} R package (version 0.7-1).}
\item L1 Random (L1R): \rev{The coefficients for L1 selected variables (i.e. variables with non--zero coefficients) are randomly permuted among the selected variables. Thus L1 is used for variable selection, but the actual non-zero coefficient estimates are replaced by noise (the coefficients of different variables). This serves as a useful benchmark (particularly in the simulations) which a well performing method should outperform.}
\end{enumerate}

% %We argue that this introduces a bias which biases the performance evaluation of the methods. This is because the $n_2$ knock out genes were selected because they are ``putative regulators'', in other words are thought to be causes of the expression of other genes.

\subsection{Performance Evaluation}\label{bias-performance}

\begin{figure}[t]
  \centering
a)  \includegraphics[width=0.4\textwidth]{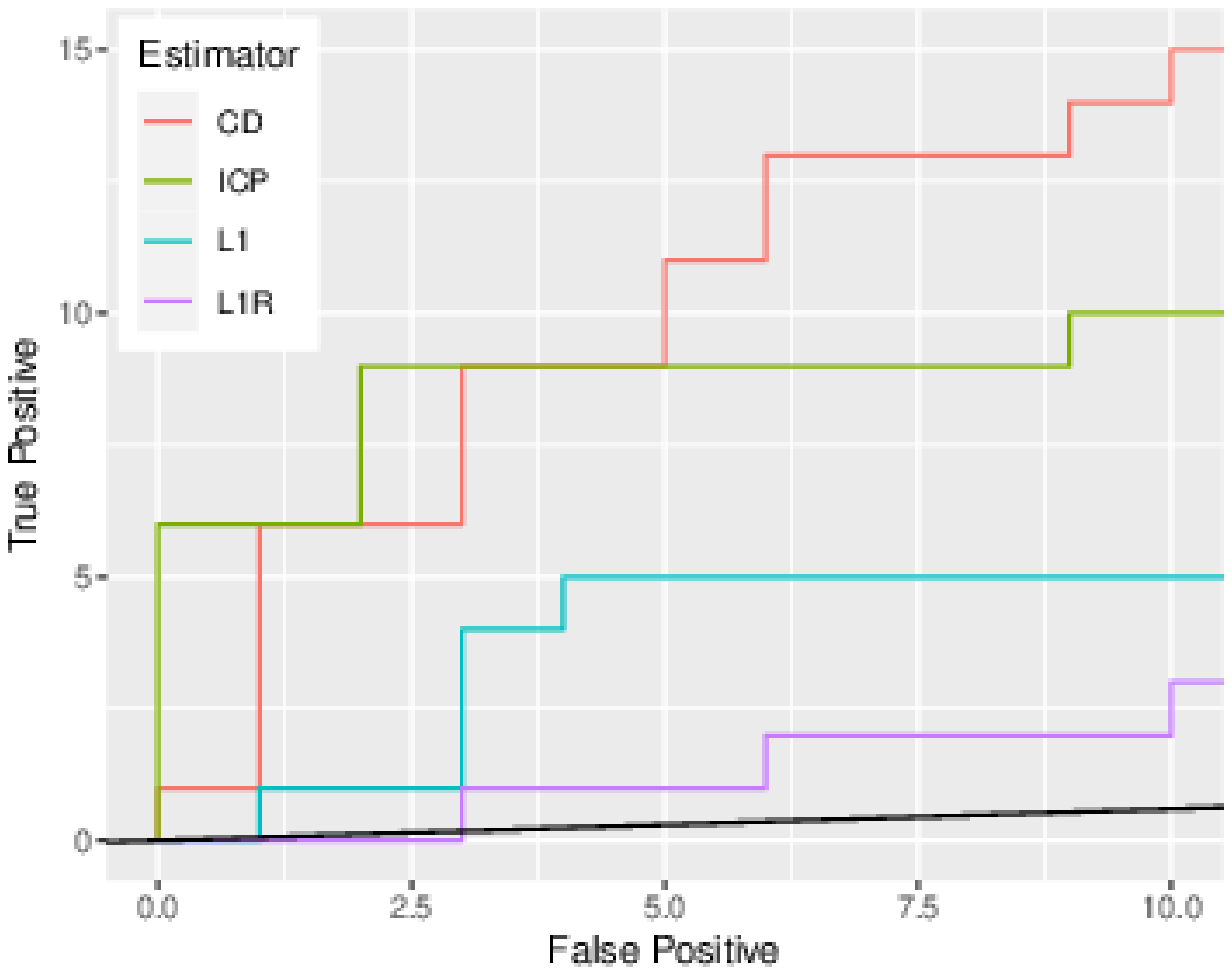}
b)  \includegraphics[width=0.4\textwidth]{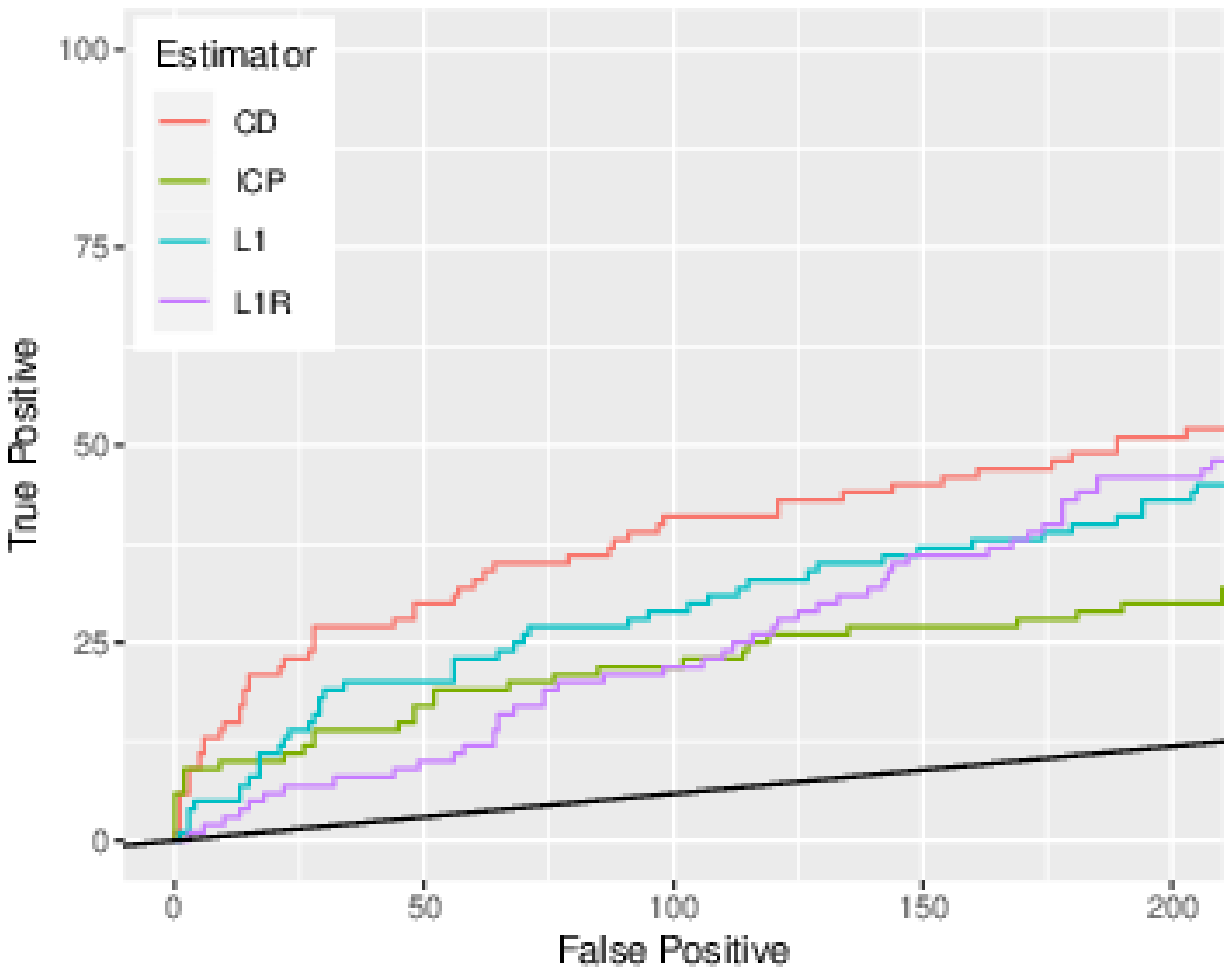}
\caption{ROC curves for estimators scoring all gene pairs with ground truth available. a) Top predictions. b) Larger set of predictions. \label{fig:roc}}
\end{figure}

Figure \ref{fig:roc} presents an ROC curve comparing the performance of the methods. Panel a) containing the top predictions is a zoomed in version of panel b). Panel a) shows that the performance of Invariant Causal Prediction (ICP) and the Causal Dantzig (CD) is particularly good for the top predictions (5/5 for ICP and 4/5 for CD). Further out on the ROC curve in b), the performance of CD, L1, and L1R are similar and ICP is worse than the other three. This roughly reproduces the results shown in Figure 2 of \cite{meinshausen2016methods} and Figure 4 of \cite{versteeg2019boosting}. The significant enrichment of causal effects in the top 5 or 10 predictions made by ICP and CD was seen in these works as evidence that these methods are identifying causal relations. For example, since only 6\% of $(i,j)$ pairs represent significant causal effects, the probably of randomly selecting 5 true significant causal effects in the top 5 predictions is less than $10^{-6}$.

To further investigate these predictions, Table \ref{tab:ICP-ranks} shows the top 10 predictions made by ICP for which ground truth was available. \rev{The column cause is the name of the purported causal gene and the column effect contains the name of the purported effect gene. The column res is TRUE if this pair is truly causal as determined by the ground truth knockout experiment and FALSE otherwise.} All of the top 5 predictions are correct and 8 of the top 10 are correct. Several of the top predicted pairs match the predictions contained in Table 4 of \cite{meinshausen2016methods}. \rev{The rank column contains the overall rank of the prediction, which includes gene pairs for which ground truth is not available. For example, the causal pair YJL141C$\rightarrow$YJL142C is the second ranked pair for which ground truth is available (because it is in second row of table) but the 8 ranked pair overall (ranks may have decimals to break ties). The pair YCL040W$\rightarrow$YCL042W is the third ranked pair for which ground truth is available (because it is in the third row) but the 17 ranked pair overall.}

\rev{The column res-flip contains ground truth for the reverse knockout. It is NA when the reverse knockout was not performed. For example, for the first row the reverse prediction is YLL020C$\rightarrow$YLL019C. Since YLL020C was not knocked out, there is no ground truth for this prediction. Thus the res-flip column is NA. Of the top 10 prediction, in only one case was the reverse knockout performed and in that case the the reverse pair was not causal (YDL074C is not a cause of YJL168C).} The column rank--flip is the rank of the flipped cause effect pair, e.g. for the first row the rank--flip is the rank of YLL020C$\rightarrow$YLL019C (which is 3). Interestingly, ICP often ranks the flipped pairs as being as likely to be cause--effect as the actual pair. For example for the first row the rank of the reverse pair YLL020C$\rightarrow$YLL019C is 3 meaning it was judged the third most likely cause effect pair of all $\sim 36$ million pairs. In two instances the flipped rank is lower than the pair itself indicating ICP believes it is more likely the flipped pair is cause--effect than the actual scored pair. \rev{Ground truth is generally not available for the often highly ranked flipped pairs, so they are not scored.}
\begin{center}
\begin{table}

\caption{ICP rankings  \label{tab:ICP-ranks}}
\centering
\begin{tabular}[t]{lllrlr}
\toprule
cause & effect & res & rank & res-flip & rank-flip\\
\midrule
YLL019C & YLL020C & TRUE & 1.0 & NA & 3\\
YJL141C & YJL142C & TRUE & 8.0 & NA & 16\\
YCL040W & YCL042W & TRUE & 17.0 & NA & 60\\
YDR432W & YDR433W & TRUE & 20.5 & NA & 7\\
YGR152C & YGR151C & TRUE & 33.0 & NA & 53\\
\addlinespace
YDR155C & YDR154C & TRUE & 52.0 & NA & 796\\
YDR101C & YLR276C & FALSE & 55.5 & NA & 28845024\\
YJL168C & YDL074C & FALSE & 61.0 & FALSE & 34506690\\
YML058W & YKL037W & TRUE & 76.0 & NA & 19112995\\
YMR104C & YMR103C & TRUE & 82.0 & NA & 14\\
\bottomrule
\end{tabular}
\end{table}

\end{center}
Thus we observe that ICP often ranks both $(i,j)$ and $(j,i)$ highly but that typically only one of these pairs is scored. There are two possible explanations for this phenomenon:
\begin{itemize}[itemsep=0pt,parsep=0pt]
\item ICP is correctly identifying feedback loops (gene $i$ is a cause of $j$ and gene $j$ is a cause of gene $i$). Under this explanation, if the reverse knockouts had been done, then the effects would alter the causes.
\item Since the Kemmeren data set obtained ground truth on ``putative regulators'', when ICP ranks both $(i,j)$ and $(j,i)$ highly, whichever is the true cause--effect pair is actually scored while the incorrect pair is not scored. \rev{Essentially if $i$ is a cause of $j$ and not the reverse, then Kemmeren is more likely to knockout $i$ (and thus obtain ground truth for $(i,j)$) than knock out $j$ and obtain ground truth for $(j,i)$.} Under this explanation, if the reverse knockouts had been done, few or none of them would have been scored as correct.
\end{itemize}
ICP is not generally consistent under feedback loops that include the target variable, a point which favors the second explanation. We note that the situation is similar, although less severe, for other methods including L1 regression and the Causal Dantzig. We conclude that the sample selection bias introduced by knocking out ``putative regulators'' may be a key driver in the performance of methods. In the following sections we seek performance metrics which are less sensitive to gene knockout selection bias and thus more informative regarding the causal predictive performance of the models.

\section{Modified Scoring Set}\label{score}

\begin{figure}[t]
  \centering
\includegraphics[width=0.6\textwidth]{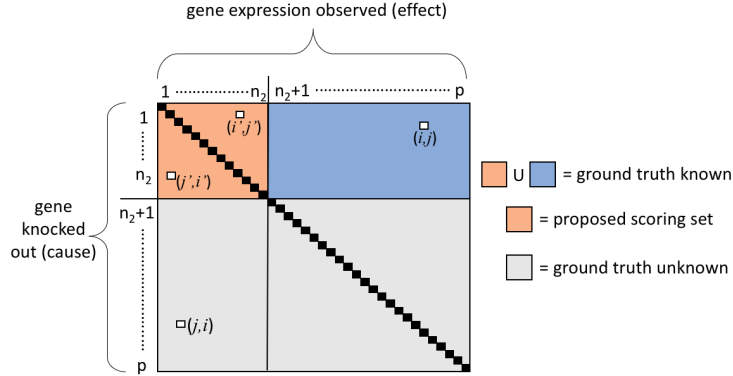}
\caption{Ground truth availability and proposed scoring set for Kemmeren. Scoring predictions on all the data where ground truth is available (union of the blue and orange regions) results in biased assessments of model performance. We propose scoring models on the orange set where ground truth for the reverse knockout is always available.\label{fig:score}}
\end{figure}

\rev{We propose scoring causal algorithms on a subset of gene pairs $(i,j)$ for which ground truth is known. We argue that this protects against the performance inflation identified in the previous section. Let}

\begin{equation*}
F_{ij} = \begin{cases}
      1 & \text{ if gene $i$ has a significant effect on gene $j$} \\
      0 & \text{ o.w. }
   \end{cases}
\end{equation*}

\rev{Causal prediction algorithms predict the $F_{ij}$ values. Since there are $p$ genes, there are $p(p-1)$ interesting $F_{ij}$ values ($F_{ii}$ is not interesting because it is the causal effect of gene $i$ on itself). Since only $n_2$ gene knockouts were performed in Kemmeren, $F_{ij}$ is known (and algorithms can be scored) on at most $n_2(p-1)$ gene pairs. Figure \ref{fig:score} illustrates the $F_{ij}$ matrix. $F_{ij}$ is known for the first $n_2$ rows (union of orange and blue regions). Scoring algorithms on this entire set can result in biased conclusions because the blue region will have a higher proportion of causal effects (higher proportion of $F_{ij}=1$) than the grey region due to selection of putative regulators as knockouts. Specifically, an algorithm can rank both $(i,j)$ and $(j,i)$ highly when $i$ and $j$ are highly correlated and then use the selection effect to only score the pair that is most likely to be causal. In the case of Figure \ref{fig:score}, $(i,j)$ will be scored and $(j,i)$ will not be scored.}

\rev{Our solution is to only score gene pairs in the orange region of Figure \ref{fig:score}. We term this scoring set $S$. $S$ satisfies the property $(i,j) \in S$ if and only if $(j,i) \in S$. Models are scored on both $(i,j)$ and $(j,i)$ or neither.  Since there are $n_2=1479$ genes which were knocked out (and the expression of every other gene was measured on these knockouts), the orange scoring region contains a total of $n_2(n_2-1)=2,185,962$ gene pairs. This represents approximately 1/16 of all pairs of genes (1/4 of genes are knocked out and these can be scored on the 1/4 of knocked out genes). Using scoring set $S$, a model cannot obtain good performance by ranking highly correlated pairs highly and letting the ``putative regulators'' selection effect score only the correct prediction. For example, if $i'$ and $j'$ are highly correlated, then an algorithm can rank both $(i',j')$ and $(j',i')$ as being causal pairs. Both of these will be scored since they are both in the orange region. In contrast, neither $(i,j)$ nor $(j,i)$ will be scored because no ground truth is available for $(j,i)$ and scoring only $(i,j)$ could bias performance measures.}

\begin{figure}[t]
  \centering
a)  \includegraphics[width=0.4\textwidth]{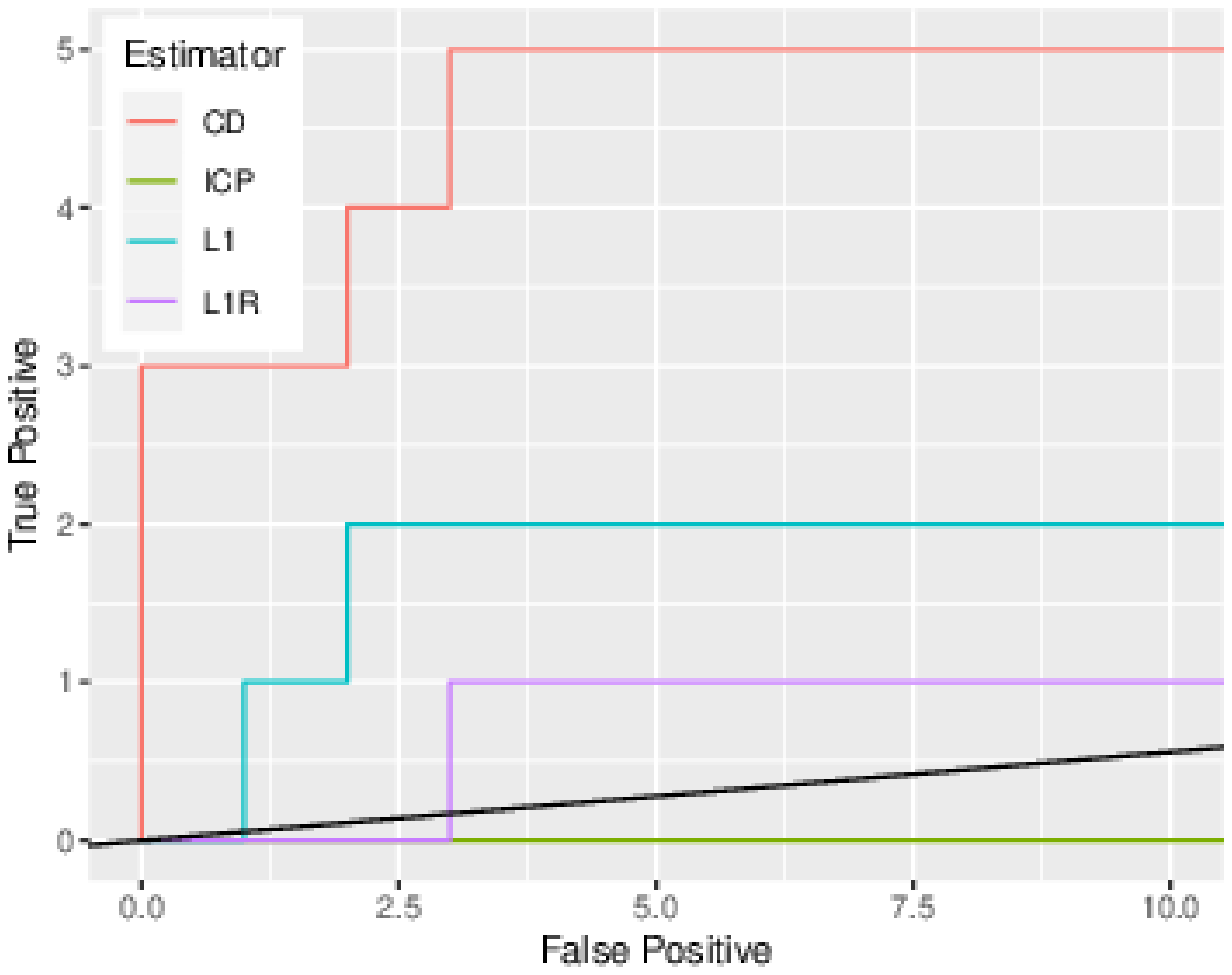}
b)  \includegraphics[width=0.4\textwidth]{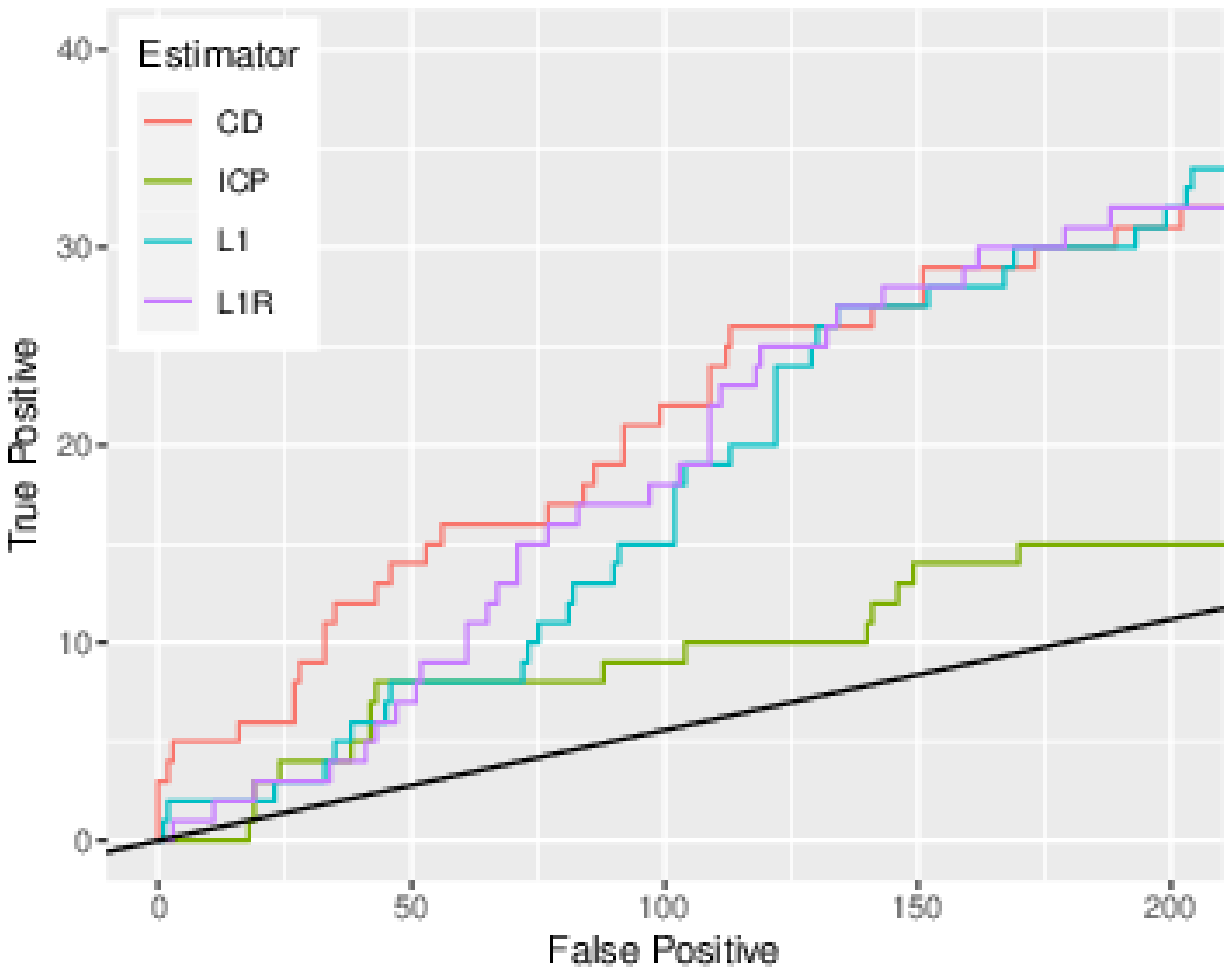}
\caption{ROC curves for estimators scoring on set $S$. a) Top predictions. b) Larger set of predictions. \label{fig:roc_sym}}
\end{figure}

\rev{In the original studies with Kemmeren, the algorithms (ICP, CD) were scored on the union of the blue and orange regions. We now score the algorithms only on set $S$ (orange region). Figure \ref{fig:roc_sym} shows the ROC curves for the four models when scoring the algorithm performance only on set $S$. The performance of all methods has significantly deteriorated. ICP is now the worst performing method. The performance improvement of CD over L1 is the result of only 3 correct predictions. There is no evidence that ICP is outperforming purely association based methods such as L1 and little evidence that CD is outperforming L1.}

\section{Simulations}\label{sim}

\rev{We test the performance of the four estimators in simulations which reproduce the structure of Kemmeren in terms of sample size ($n$), number of variables ($p$), number of knockouts performed ($n_2$), and type of knockout (single gene knockouts with each gene knocked out at most once). The simulations use linear models without any hidden confounding or causal feedback loops. This setting is designed to be favorable for ICP and CD because both estimators assume linear models and ICP assumes no hidden confounding.}

\rev{A main goal of these simulations is to study when and by how much causal estimators such as CD and ICP can outperform association based methods such as L1 regression. Towards this end, we simulate under parameter settings which are easy for L1 and very difficult for L1. In the former, one hopes that CD and ICP can match the performance of L1 while in the latter CD and ICP can beat L1.}

\subsection{Simulation Parameters}

\rev{Data is generated from a directed acyclic graph with $p+1$ nodes $X_1,\ldots,X_{p+1}$ where $p=6400$. Let upper triangular matrix $A$ represent edges in the DAG with $A_{ij}$ the causal effect of $X_i$ on $X_j$. The response of interest is $Y=X_{p/2+1}$. The goal is to identify direct causes of $Y$ i.e. determine which elements of $A_{\cdot,p/2+1}$ ($p/2+1$ column of $A$) are non--zero.}

%We consider several connection densities. Let $p_c$ be the probability that $X_i$ and $X_j$ are directly connected. We consider $p_c=(0.003125,0.00625,0.0125,0.025,0.05)$. This implies that $X_1$ is directly connected with an expected number of $20,40,80,160,320$ nodes.

%We let $p_0$ be the number of true direct causes (parents) of $Y$ and consider $p_0=1$ and $p_0=2$. In addition there are $p_0$ direct effects (children) of $Y$. These $p_0$ direct effects of $Y$ can challenge association based estimators because they may be more strongly correlated with $Y$ than the causes of $Y$.

The matrix $A$ is randomly generated on each simulation run. We simulate Strong causes where the causal coefficients of $X$ on $Y$ are large (relative to the effects of $Y$) and Weak causes where they are small (again relative to the effects of $Y$). \rev{In the Strong causes case we expect L1 regression to perform well because the nodes which are most strongly associated with $Y$ are actually causes of $Y$. In the Weak case, we expect L1 regression to perform poorly because the nodes which are most strongly associated with $Y$ are actually effects of $Y$.} $A$ is generated in the following manner:
\begin{itemize}
[itemsep=0pt,parsep=0pt]
\item $A_{ij}$ is drawn independently with $P(A_{ij}=-1) = P(A_{ij}=1) = n_t/(2p)$ and $P(A_{ij}=0) = 1-n_t/p$. The $n_t$ parameter controls the density of the connections in the network. We consider $n_t = 20,40,80,160,320$. \rev{For the $n_t=20$ simulation, $X_1$ has approximately $20$ descendants and for the $n_t=320$ simulations, $X_1$ has approximately $320$ descendants.}
\item The lower triangle (and diagonal) entries of $A$ are subsequently set to $0$, ensuring $A$ represents an acyclic graph.
\item All causes and effects of $Y$ are set to $0$ ($p/2+1$ row and column of $A$ set to $0$).
\item Nodes $X_{p/2+1-p_0},\ldots,X_{p/2}$ are set to be causes of $Y=X_{p/2+1}$ and nodes $X_{p/2+2},\ldots,X_{p/2+p0}$ are set to be effects of $Y$. Note that since $p_0$ is $1$ or $2$, this amounts to $Y$ having either 1 direct cause and 1 direct effect ($p_0=1$ case) or 2 direct causes and 2 direct effects ($p_0=2$ case). Specifically $P(A_{k,p/2+1}=s_1) = P(A_{k,p/2+1}=-s_1) = 1/2$ for $k=p/2+1-p_0,\ldots,p/2$ and $P(A_{p/2+1,k}=s_2) = P(A_{p/2+1,k}=-s_2) = 1/2$ for $k=p/2+2,\ldots,p/2+1+p_0$. The parameters $s_1$ and $s_2$ control the strength of the causes of $Y$ and the effects of $Y$ respectively. Large values of $s_1$ result in causes of $Y$ which are highly correlated with $Y$. Similarly large values of $s_2$ result in the effects of $Y$ being strongly correlated with $Y$. Thus the regime where $s_1$ is small and $s_2$ is large is challenging for association based estimators because the $X$ nodes which are most strongly correlated with $Y$ are the effects of $Y$, not the causes of $Y$. In the Strong (easy for L1) setting, $s_1 = s_2 = 1$ and in the Weak (challenging for L1) setting, $s_1 = 1$ and $s_2 = 1000$.
\item $A$ is column normalized.
\end{itemize}

In each simulation run, $n_1=300$ non-interventional observations are generated and $n_2=p/4$ knockouts are generated. Thus there are a total of $n=1900$ observations. Following Kemmeren, each gene is knocked out either $0$ or $1$ times. Non--interventional data is generated according to $X_1 \sim \epsilon_1$ and
\begin{equation*}
X_j = \sum_{i=1}^{j-1} A_{ij}X_i + \epsilon_j
\end{equation*}
for $j > 1$ where $\epsilon_j \sim N(0,1)$. For the interventional data, when gene $j$ is knocked out its expression is shifted by $-40$. The $p_0$ direct causes and $p_0$ direct effects of $Y$ are never knocked out.

\subsection{Fitting Models and Performance Evaluation}

\rev{The four models (ICP, CD, L1, and L1R) are fit as described in Section \ref{kemmeren}. Performance is evaluated in the following manner: Each model returns a coefficient vector $\widehat{\beta}$ where $\widehat{\beta}_j$ is the estimated causal effect of $X_j$ on $Y$.} The estimates are ranked by absolute size and the number of true causes among the top ranked $p_0$ estimates is recorded. For $p_0=1$ the number of true causes among the top ranked $p_0$ estimates is either $0$ or $1$. For $p_0=2$, it is $0,1$ or $2$. We compute the mean number of true causes in the top ranked $p_0$ estimates across the $N=100$ simulation runs, regenerating $A$, $\epsilon_X$, and the set of genes to knockout on each run. The best possible performance for the $p_0=1$ simulations is $1$: In all $N$ simulation runs the true cause of $Y$ always had the largest parameter coefficients in absolute size. The best possible performance for the $p_0=2$ simulations is $2$: In all $N$ simulation runs the two true causes of $Y$ coefficients were always the largest two parameter estimates in absolute size. We compare the performance of the same four methods analyzed in Section \ref{kemmeren}. We do not implement the bootstrap sampling strategy for these results.

On a given simulation run, ICP and/or CD may return all $0$ parameter estimates or an error. For example, the maximin coefficients computed by ICP are $0$ when the model does not believe there is sufficient information to assign causality to any variable. The CD requires the inverse of the difference in two Gram matrices to be well--defined. If this difference is not invertible, the CD estimator is undefined. In these case, we assign these methods the L1R parameter estimates. This is most likely to happen when the interventions are not sufficiently complex to estimate the causal parameters \citep{rothenhausler2019}.

\subsection{Results}

\begin{center}
\begin{table}

\caption{Simulations results. For the $p_0=1$ rows, the best performance is 1 and the worst performance is 0. For the $p_0=2$ rows, the best performance is 2 and the worst performance is 0. \label{tab:high-dim}}
\centering
\begin{tabular}[t]{rrrrrrrrr}
\toprule
\multicolumn{1}{c}{ } & \multicolumn{4}{c}{Strong} & \multicolumn{4}{c}{Weak} \\
\cmidrule(l{3pt}r{3pt}){2-5} \cmidrule(l{3pt}r{3pt}){6-9}
$n_t$ & L1 & L1R & CD & ICP & L1 & L1R & CD & ICP\\
\midrule
\addlinespace[0.3em]
\multicolumn{9}{l}{\textbf{$p_0=1$}}\\
\hspace{1em}20 & 1 & 0.27 & 0.70 & 0.29 & 0 & 0.19 & 0.43 & 0.19\\
\hspace{1em}40 & 1 & 0.32 & 0.71 & 0.34 & 0 & 0.34 & 0.37 & 0.34\\
\hspace{1em}80 & 1 & 0.25 & 0.82 & 0.27 & 0 & 0.25 & 0.35 & 0.25\\
\hspace{1em}160 & 1 & 0.32 & 0.84 & 0.33 & 0 & 0.26 & 0.46 & 0.26\\
\hspace{1em}320 & 1 & 0.28 & 0.85 & 0.30 & 0 & 0.32 & 0.39 & 0.32\\
\addlinespace[0.3em]
\multicolumn{9}{l}{\textbf{$p_0=2$}}\\
\hspace{1em}20 & 2 & 0.96 & 1.37 & 0.97 & 0 & 0.95 & 0.78 & 0.95\\
\hspace{1em}40 & 2 & 1.02 & 1.47 & 1.02 & 0 & 1.05 & 0.79 & 1.05\\
\hspace{1em}80 & 2 & 1.06 & 1.52 & 1.06 & 0 & 0.99 & 0.66 & 0.99\\
\hspace{1em}160 & 2 & 1.16 & 1.61 & 1.17 & 0 & 1.00 & 0.78 & 1.00\\
\hspace{1em}320 & 2 & 1.10 & 1.77 & 1.11 & 0 & 1.01 & 0.82 & 1.01\\
\bottomrule
\end{tabular}
\end{table}

\end{center}

Results are summarized in Table \ref{tab:high-dim}. With Strong causes, L1 regression obtains perfect performance, $1$ for $p_0=1$ case and $2$ for $p_0=2$ case. This is not surprising. In the Strong case the true cause of $Y$ is very highly correlated with $Y$ relative to all other variables and so has the largest parameter coefficient.

\rev{Recall that L1R randomly permutes the ranks of coefficients selected by L1. Thus a method which matches the performance of L1R is providing no performance benefit beyond what is already provided by the L1 variable prescreening.} L1R obtains performance of approximately $0.25$ for the $p_0=1$ case and $1$ for the $p_0=2$ case. This is expected. Lasso regression selects a model with 4 variables. This will typically include the true causes (1 if $p_0=1$ and 2 if $p_0=2$) of $Y$, along with 3 (if $p_0=1$) or 2 (if $p_0=2$) non--causes of $Y$. Thus when randomly permuting coefficients, L1R will select the true coefficient about 25\% of the time when $p_0=1$ and will have on average 1 correct in the top 2 when $p_0=2$. 

\rev{The CD has performance worse than L1 in all Strong settings tested. Relatively speaking CD performs best with dense networks (high $n_t$) values. For example, in the most favorable setting for CD relative to L1 ($p_0=2$ and $n_t=320$), CD identifies about 20\% fewer causal effects than L1. This can be explained by the fact that with sparser networks (such as $n_t=20$ where each node effects few other nodes), very few of gene knockouts will effect the true causes and effects of $Y$. Thus the perturbation environment distribution will resemble the observational environment data distribution, a challenging setting for CD. ICP has worse performance than CD and is comparable to L1R.}

The Weak columns represent a very challenging setting for L1. The effects of $Y$ are consistently more strongly correlated with $Y$ than the causes of $Y$. \rev{Thus the top ranked genes by L1 are expected to be effects of $Y$, not causes.} At all $p$ and $p_0$, L1 never selects the true causes of $Y$ in its top 1 or 2 predictions. This is the worst possible performance a method can have. Here is a setting where we may hope a causal estimator can outperform L1.

Indeed both CD and ICP perform consistently higher than L1. In this case, L1R provides a useful performance benchmark for CD and ICP. ICP performs very similarly to L1R. This is because the maximin coefficients returned by ICP are very often $0$, so the method defaults to the L1R parameter estimates. Qualitatively ICP is returning $0$ maximin parameter estimates because it is unable to determine causes of $Y$. CD has somewhat better performance than L1R for $p_0=1$ and somewhat worse performance than L1R for $p_0=2$.

Our general conclusion is that causal inference with sample sizes approximating Kemmeren is quite difficult. Under the scenarios considered CD and ICP are unable to consistently beat L1 and L1R models, both association based estimators. \rev{ICP has particularly poor performance.} While these simulations do not reproduce all the complexity of an actual genetic perturbation experiment, they provide a setting to test the models which is not subject to sample selection bias.

\hypertarget{discuss}{%
\section{Discussion}\label{discuss}}

Sample selection bias is a well known issue in non--causal prediction models \citep{huang2006correcting,globerson2006nightmare}. \rev{In this work we demonstrated that sample selection bias in Kemmeren is likely a key driver of the reported predictive performance of several recently proposed causal estimators.} To avoid potential sample selection bias, we suggest that future studies with Kemmeren examine the ranks of flipped predictions (as done in Table \ref{tab:ICP-ranks} here), evaluate prediction performance on the set $S$ (as proposed in Section \ref{score} here), and conduct simulations studies which can be used to evaluate models in a way that is free of sample selection bias. More broadly, the generation and curation of data sets suitable for causal prediction will aid in the unbiased evaluation of causal predictive models.

\section*{Acknowledgements}

J.P.L. was supported by NIH/NCI grants P50CA127001, P30CA016672, and P50CA140388. M.J.H. was supported by NIH/NCI grants R21CA22029 and 1R01CA244845-01A1.

%\nocite{*}% Show all bib entries - both cited and uncited; comment this line to view only cited bib entries;
\bibliographystyle{abbrvnat}
\bibliography{refs}%

%\section*{Author Biography}

%\begin{biography}{\includegraphics[width=60pt,height=70pt,draft]{empty}}{\textbf{Author Name.} This is sample author biography text this is sample author biography text this is sample author biography text this is sample author biography text this is sample author biography text this is sample author biography text this is sample author biography text this is sample author biography text this is sample author biography text this is sample author biography text this is sample author biography text this is sample author biography text this is sample author biography text this is sample author biography text this is sample author biography text this is sample author biography text this is sample author biography text this is sample author biography text this is sample author biography text this is sample author biography text this is sample author biography text.}
%\end{biography}

\end{document}